\documentclass{article}
\usepackage[preprint]{nips_2018}

\usepackage{algorithm}
\usepackage{amsfonts}		
\usepackage{amsmath}
\usepackage{booktabs}		
\usepackage{color}
\usepackage{comment}
\usepackage{etoolbox}
\usepackage[T1]{fontenc}	
\usepackage{graphicx}
\usepackage{hyperref}		
\usepackage[utf8]{inputenc}	
\usepackage{microtype}		
\usepackage{nicefrac}		
\usepackage{url}			

\newbool{inccomment}
\setbool{inccomment}{true}
\newcommand{\XX}[1]{\ifbool{inccomment}{{\color{magenta} #1}}{}}
\newcommand{\CT}[1]{\ifbool{inccomment}{{\color{magenta}CT\@: #1}}{}}
\newcommand{\NT}[1]{\ifbool{inccomment}{{\color{blue}NT\@: #1}}{}}
\newcommand{\TD}[1]{\ifbool{inccomment}{{\color{orange}#1}}{}}
\newcommand{\FN}[1]{\ifbool{inccomment}{{\color{OliveGreen}#1}}{}}
\newcommand{\GR}[1]{\ifbool{inccomment}{{\color{Tan}#1}}{}}
\newcommand{\LD}{\ifbool{inccomment}{{\color{magenta}\\============================================\\}}}
\newcommand{\RF}{\ifbool{inccomment}{{\color{green}~[R]}}}

\newcommand{\roma}[1]{\uppercase\expandafter{\romannumeral #1\relax}}

\title{Towards Robust Training of Neural Networks\\ by Regularizing Adversarial Gradients}

\author{
	Fuxun Yu$^1$, Zirui Xu$^2$, Yanzhi Wang$^4$, Chenchen Liu$^5$, Xiang Chen$^3$\\
	$^{1,2,3}$Department of Electrical Computer Engineering, George Mason University, Fairfax, VA 22030 \\
	$^{4}$Department of Electrical Computer Engineering, Northeastern University, Boston, MA 02115 \\
	$^{5}$Department of Electrical Computer Engineering, Clarkson University, Potsdam, NY 13699 \\
	\texttt{fyu2@gmu.edu$^1$, zxu21@gmu.com$^2$, yanzhiwang@northeastern.edu$^4$ }\\ \texttt{chliu@clarkson.edu$^5$, xchen26@gmu.com$^3$} \\
}

\begin{document}
\maketitle

\begin{abstract}
In recent years, neural networks have demonstrated outstanding effectiveness in a large amount of applications.
	However, recent works have shown that neural networks are susceptible to adversarial examples, indicating possible flaws intrinsic to the network structures.
	To address this problem and improve the robustness of neural networks, we investigate the fundamental mechanisms behind adversarial examples and propose a novel robust training method via regulating adversarial gradients.
	The regulation effectively squeezes the adversarial gradients of neural networks and significantly increases the difficulty of adversarial example generation.
	Without any adversarial example involved, the robust training method could generate naturally robust networks, which are near-immune to various types of adversarial examples.
	Experiments show the naturally robust networks can achieve optimal accuracy against Fast Gradient Sign Method (FGSM) and C\&W attacks on MNIST, Cifar10, and Google Speech Command dataset.
	Moreover, our proposed method also provides neural networks with consistent robustness against transferable attacks.
\end{abstract}

\section{Introduction}
\label{sec:intro}
\vspace{-1mm}
Recently, researchers have found that neural networks are vulnerable to adversarial examples, which are natural inputs polluted with small-magnitude, additive perturbations [1,2].
	Adversarial examples strongly influence classification accuracy with comparatively small changes to the underlying data.
	In this paper, we alleviate the effect of adversarial examples and improve neural networks' robustness by considering neural networks from a quasi-linear point of view.

Suppose a neural network could be seen as a huge function $F_{\theta}(x)$ with loss function $L_{\theta}(x)$ and network parameter set $\theta$.
	We can linearly approximate the loss function by its first-order Taylor expansion at point $x$:
\begin{equation}
	L_{\theta}(x+\Delta x) = L_{\theta}(x) + \frac{\partial L_{\theta}(x)}{\partial x} \cdot \Delta x,
	\label{eq:1}
\end{equation}

where $\Delta x$ means a unit change of input $x$ in its neighborhood, ${\partial L_{\theta}(x)}/{\partial x}$ is the derivative of $L_{\theta}(x)$ at point $x$, namely the gradients.
	During adversarial attacks, large first-order gradients might be leveraged to amplify small adversarial perturbation $\Delta x$ and mislead the classification output eventually.
	The existence of those leveraged gradients (\textit{i.e.} adversarial gradients) reflects the network vulnerability to adversarial examples.
This is essentially where we target and propose to tackle the adversarial attack problem:
\emph{by regulating the adversarial gradients to prevent small-magnitude perturbation from exerting large influence.}



To accomplish this, we introduce the concept of gradient regularizer to enhance the network robustness, which forces the adversarial gradients to shrink and eliminates the large influence from small magnitude adversarial perturbations.
	The gradient regularizer is also designed to discriminate adversarial gradients from normal ones, preventing overall accuracy drop caused by undifferentiated gradient decay.
	The major contributions of our robust training method are as:
	
	\hspace{4mm} $\bullet$ We introduce gradient loss function into network training for regulating adversarial gradients.
		And a dedicated gradient regularizer is designed to precisely capture the suspicious adversarial gradients, utilizing logit layer's outputs and label information;
	\vspace{-1mm}
	
	\hspace{4mm} $\bullet$ We adopt double-backpropagation [3,4] to process the second-order gradients in the gradient loss function, which enables simultaneous gradient regulation in addition to network training;
	\vspace{-1mm}

	\hspace{4mm} $\bullet$ We propose a comprehensive robust training process based on adversarial gradient regularization and cross-entropy based network training.
		Such a robust training could significantly enhance network's resistance against various adversarial attacks, \textit{e.g.} FGSM [2], C\&W attack [5], \textit{etc}.;
	\vspace{-1mm}
	
	\hspace{4mm} $\bullet$ We further explore the potential of the proposed robust training beyond aforementioned white-box attacks, and verify its optimal resistance against transferable attacks.

Experiments show that, through our proposed gradient regularized training, we can significantly increase the accuracy against FGSM generated adversarial example to 30\% $\sim$ 91\% on MNIST, 26\% $\sim$ 53\% on CIFAR10 and 16\%$\sim$ 55\% on Google Speech Command dataset.
	Against the current strongest C\&W attack, our training method could recover the accuracy from zero back to 32\% $\sim$ 92\% on MNIST, 17\% $\sim$ 48\% on CIFAR10, and 9\% $\sim$ 48\% on Google Speech Command dataset.
	Besides white-box attacks, our robust trained models demonstrate much stronger defensive ability against transferable attacks, providing at most 50\%, 80\% and 40\% accuracy on MNIST, CIFAR10 and Google Speech Command dataset. 
	
According to experimental results, our robust training method outperforms currently state-of-art gradient regulating method [6] under nearly all adversarial settings. 
In terms of training overhead, our method only introduce one extra backpropagation overhead per training step. 
	Compared to Min-Max optimization [15], which needs to generate over tons of adversarial examples, our method has great advantages regarding robustness boost together with time or resource consumption of model training. Therefore, our gradient regularizer can be deployed on various existing neural networks by fine-tuning a few epochs by our method, allowing it to be utilized as a common method to enhance the robustness of many current widely-used neural networks, AlexNet [7], VGG [8], \textit{etc}.
\section{Related Work}
\label{sec:related}

Adversarial examples were first introduced in [2], where the authors generated adversarial examples by solving the box-constrained optimization problem.
	Later, Goodfellow \textit{et al.} proposed the Fast Gradient Sign Method (FGSM) [2], which could generate untargeted adversarial examples by calling backpropagation only once.
	This method could also be repeated by multiple times, resulting in a Basic Iterative Method [2] for either targeted or untargeted attacks.
	Alternately, the Jacobian-based Saliency Map Approach (JSMA) [9] used a greedy algorithm that could identify the most influential pixels by calculating the adversarial saliency map, then perturbing those top ranked pixels iteratively.
	Moreover, C\&W attacks [5] are currently considered as the most effective adversarial attacks, which are a series of $\ell_0$, $\ell_1$, $\ell_2$, and $\ell_\infty$ attacks with a stronger adversarial loss function.
	Recently, more attacks emerged based on network interpretability, such as transferable attack [10,11]. It aims to produce highly transferable adversarial examples that can fool multiple neural networks with different structures and parameters [12].

Current defense techniques include adversarial training [2], defensive distillation [13,14], Min-Max robustness optimization [15], \textit{etc}.
	Adversarial training techniques augment the natural training samples with corresponding adversarial examples together with correct labels.
	Defensive distillation [13] aggressively boost the classification confidence gap between correct and wrong predictions to increase the manipulation difficult for adversarial examples.
	The recently proposed Min-Max robustness optimization [15] aims to augment the training dataset with large amount of adversarial examples within a $l_\infty$-norm ball to compensate the largest loss decrease.

However, these methods also have certain shortcomings.
	In adversarial training, the robustness improvement hugely relies on the existing adversarial examples in the training set and cannot generalize well on unseen adversarial examples.
	Meanwhile, Min-Max optimization needs to generate adversarial sample over ten times more than original dataset size, which is extremely time-consuming and inapplicable to big dataset (\textit{e.g.} ImageNet).
	As for defensive distillation, attackers could use logits layer output to obtain adversarial gradients, e.g. C\&W attack [5], thus bypassing the SoftMax layer temperature and defeating it.

Compared to these methods, our robust training method improves model's robustness by regulating gradients during natural training process. 
	Therefore, there is no need to generate any adversarial example, ensuring minimal overhead compared to adversarial training and Min-Max optimization. 
	In addition, our gradient regularizer is designed based on the logit layer output, which could provide consistent robustness against various of adversarial attacks, \textit{e.g.} FGSM [2], C\&W attack [5], \textit{etc}.
	In the next section, we will delve into the details of our proposed robust training method.


\section{Robust Training by Regularizing Adversarial Gradients}
\label{sec:robust}

\begin{figure}[b]
  \centering
  \includegraphics[width=5.5in]{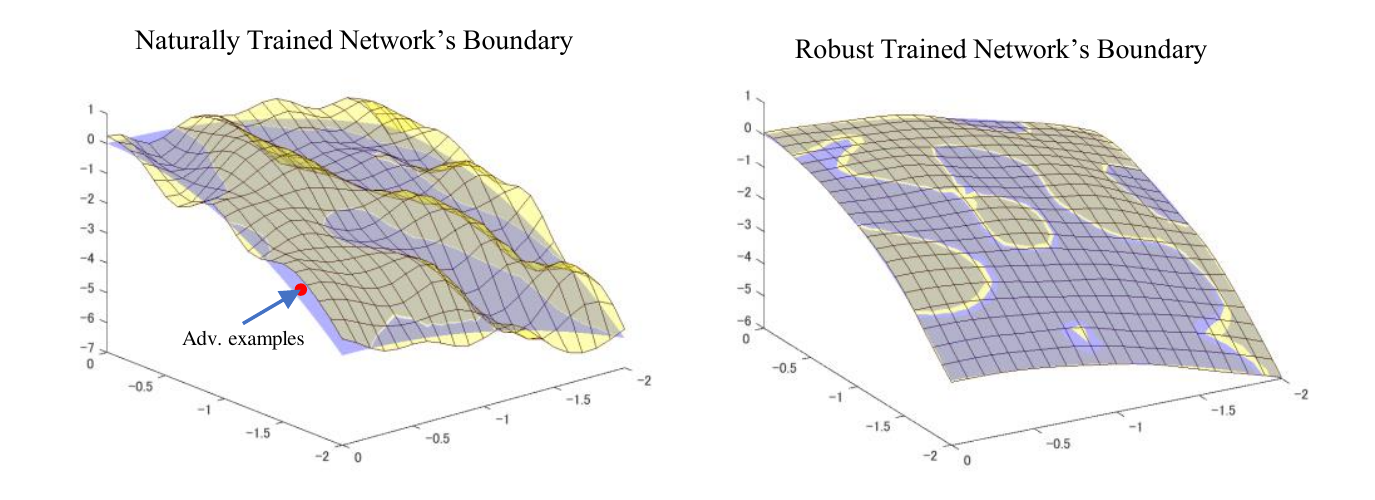}
  \caption{The illustration of neural network's learned boundary (yellow) and the correct boundary (blue). Naturally trained neural networks' decision boundary is usually full of peaks and valleys. By contrast, network trained by combined cross-entropy and gradient loss could form a relatively smooth boundary without large slopes, and thus is more resistant to small adversarial perturbations.}
  \label{fig:1}
\end{figure}

In this section, we first formulate the loss function of robust training by taking gradient loss into consideration.
	Then a dedicated gradient regularizer is proposed as an implementation of the gradient loss function to discriminate and penalize adversarial gradients.
	To process the second-order gradients in the gradient regularizer, we further integrate double-backpropagation in our computing process.
	Finally, we propose a comprehensive robust training process based on adversarial gradient regularization and cross-entropy based network training.

\paragraph{Robust Training Loss Formulation}
As aforementioned, large first-order gradients might be leveraged to amplify small adversarial perturbations and mislead the classification output.
	Therefore, we intuitively enforce these adversarial gradients to be as small as possible while maintaining the same accuracy.
	As loss function could be used in training process to update the network parameters to satisfy certain constraints, we introduce the gradient loss $L_{grad}$ into the original cross-entropy loss function.
	With regulation of the loss functions, the large gradients would be effective penalized in the network training process with optimal network accuracy, preventing potential adversarial attacks.
	Our loss function could be formulated as follows:
\begin{equation}
  \begin{aligned}
    & L_{\theta}(x) = L_{ce} ~+~ c~\cdot~L_{grad}, \\
    whe&re ~~ L_{grad} = L_{norm} (\frac {\partial Reg_{\theta}(x)}{\partial x}),
    \label{eq:2}
\end{aligned}
\end{equation}
where $L_{ce}$ is normal cross-entropy loss, $L_{grad}$ is gradient loss, $L_{norm}$ is the gradients matrix and ${\partial Reg_{\theta}(x)}/{\partial x}$ is the Jacobian matrix of our regularizer function $Reg_{\theta}(x)$.
	The coefficient $c$ here is used to adjust the gradient loss penalty strength so that it won't harm the training process.

Fig.~\ref{fig:1} illustrates two decision boundaries of networks trained by un-regularized cross-entropy loss function and our combined loss function respectively. As it shows, the natural trained model are more likely to learn a decision boundary with much steeper peaks and valleys. This is because in the network training process, naturally trained network are more likely to form large gradients since there is no penalty on such gradients. 
While in our robust trained model, the large gradients will be penalized so that the final formed curve is more smooth than the naturally trained model. 
When facing possible adversarial attacks, the natural trained network's large gradients are more likely to be utilized by the attacker to generate an adversarial example and lead to a misclassification. By contrast, it's hard to use the same magnitude perturbation to mislead our robust trained model, since its average gradients are much smaller.

\paragraph{Gradient Regularizer Design}
The robust training loss is formulated by regulating large gradients, which are extremely vulnerable adversarial perturbation.
	However, in practice it is critical to precisely discriminate adversarial gradients, preventing overall accuracy drop caused by undifferentiated gradient decay.
	Such a critical performance is mainly determined by the gradient regularizer design.

In previous works [6], cross-entropy is directly used as the gradient regularizer function.
	However, cross-entropy as regularizer function will imposes gradient penalty on all its input gradients without any differentiation, which can interference the gradient descent process of cross-entropy loss. 
	Another shortcoming is that, wrong label logits' coefficients are all set to zero according to cross-entropy's definition, such regularizer cannot provide constraints on other wrong logit outputs change. 

	To address these problems, we propose a novel regularizer function composed of logits layer output and input label information. Compared to probability output in softmax layer, we use the logit outputs that haven't been normalized (softmax) or discarded (cross-entropy), thus could provide more direct information to help distinguish the adversarial gradients.Our regularizer is as follows:
\begin{equation}
  Reg_{\theta}(x) = max\{ Z(x)_i,~ i \neq t\}~-~Z(x)_t,
  \label{eq:3}
\end{equation}
where $Z(\cdot)$ is the logits output before the softmax layer, and $t$ is the input $x$'s correct label.

Our regularizer function $Reg_{\theta}(x)$ could be interpreted as the difference between the maximum wrong and correct logits.
The gradient of this regularizer function means the fastest direction we could push the wrong logits to be larger than the correct ones, \textit{i.e.} causing misclassification of neural networks.
	Therefore, we consider this regularizer as the most effective regularizer since it provides the most effective adversarial gradients to cause misclassification.
	As long as we could limit the gradients of this function to be near zero, any small magnitude perturbation cannot be able to change the classification results.

\paragraph{Double-Backpropagation for Second-Order Gradients Processing}

Different from normal gradient descent problems by using first-order gradients, introducing the gradient loss into the training loss needs us to solve a second-order gradient computing problem. To solve this problem, we adopt \textit{Double-Backpropagation} [5] technique.

Applying \textit{Double-Backpropagation} to our problem, we could first compute the cross-entropy and gradient loss by forward-propagation, with the gradients then being calculated by backpropagation. Then, to minimize the gradient loss, we need to calculate the second-order derivative of gradient loss. Therefore, a second backpropagation operation is performed to compute the second-order derivative of $L_{grad}$ on $\theta$. After this, the weights of neural networks are finally updated according to gradient descent algorithm:
\begin{equation}
  \theta~' = \theta - (\frac {\partial L_{ce}}{\partial \theta}) - (\frac {\partial L_{norm}(\partial Reg_{\theta}(x) / \partial x)}{\partial \theta ~}).
  \label{eq:4}
\end{equation}
Here, $-(\frac {\partial L_{ce}}{\partial \theta})$ is the first-order gradients to minimize the cross-entropy, and $- (\frac {\partial L_{norm}(\partial Reg_{\theta}(x) / \partial x)}{\partial \theta ~})$ is the second-order partial derivative to minimize the first-order gradients of $Reg_{\theta}(x)$.

\paragraph{Robust Training Overview}
In summary, our robust training method introduces a new gradient loss in the normal training procedure to regulate the adversarial gradients, which are differentiated by our designed regularizer function $Reg_{\theta}(x)$. Except that double back-propagation costs one more back-propagation, no other training overhead is introduced into our training process. Therefore, we could maintain the convergence speed at the same order as normal training procedure while significantly enhance the model's robustness, which is far more efficient than most of adversarial training based method, e.g. Min-Max optimization [15], \textit{etc}.

In the next section, we evaluate our robust training method by training models against various of adversarial attacks and compare our results with current state-of-art gradient regularizing method.

\section{Experimental Results}

In this section, we evaluate our robust training method on both image classification and speech command classification tasks. 
Among them, MNIST and CIFAR10 dataset are selected in the image classification task, while Google speech command dataset is used in the speech command classification task.
We test all the proposed training models against three different adversarial attacks: Fast Gradient Sign Method (FGSM) attack, Basic Iterative Method (BIM) attack with cross-entropy loss, and C\&W attack with C\&W loss.
Different with the FGSM which is a one-step attack,  BIM and C\&W are iterative attacking methods and related testing parameters are set as the default values without any specification: 10 iterations with step size of 0.05, C\&W loss confidence factor $k$ = 50.
Three scenarios, namely, \emph{Natural} model, \emph{Xent-reg} model, and \emph{Logit-reg} model are evaluated and compared. 
Natural model refers to naturally trained model without using any defensive techniques, 
Xent-reg model indicates the trained model with the cross-entropy gradient regularized [6], and Logit-reg model is trained by our robust training method.


\subsection{Robust Training Performance on MNIST}

In the evaluation on MNIST dataset, a four-layer neural network model [15] with two convolutional layers and two fully connected layers is adopted, and 99.17\% accuracy is achieved based on vanilla training. 
The experimental results of the three training models, i.e. natural, Xent-reg, and Logit-reg model on the MNIST are described in Table.~\ref{table-mnist}.
We evaluate the performance under different attacking strength, which is decided by the parameter of $l_{\infty}$-norm and is set to be 0.1, 0.2, and 0.3 in this work.
The results show that the Logit-reg model with our proposed training can achieve over 90\% accuracy against all the three adversarial attacks with $l_{\infty} = 0.1$.
Compared with the Xent-reg model [6], in defending FGSM attack, our method can achieve comparable accuracy under $l_{\infty} = 0.1$ and outperform 13.4\% (54.1\%) under $l_{\infty} = 0.2$ ($l_{\infty} = 0.3$). 
With the increasing of the attack strength (i.e. $l_{\infty}$), the effectiveness of Xent-reg model [6] decreases dramatically and can hardly be used, esp. under the BIM and C\&W attack.
However, our proposed method can still perform effective defense.
For example, under C\&W attack and $l_{\infty} = 0.2$, Xent-reg model [6] obtains only 20\% accuracy, while our method can achieve 69.4\% accuracy.
Overall, under $l_{\infty} = 0.2$, our proposed Logit-reg model achieves about 3$\times$ accuracy in defending BIM and C\&W attack, compared with the current state-of-art cross-entropy gradient regularizing method (i.e. Xent-reg model).
Note that the Xent-reg fails to defend the C\&W attack totally when the $l_{\infty}$ increases to 0.3, however, our Logit-reg model can still offer 32.4\% accuracy. 




\begin{table}
  \caption{Test Accuracy of adversarial examples on MNIST dataset (\%)}
  \label{table-mnist}
  \centering
  \begin{tabular}{lllllllllll}
    \toprule
    \multicolumn{3}{}{}     & FGSM   &&&    BIM       &&& C\&W   \\
    \cmidrule(r){3-5} \cmidrule(r){6-8} \cmidrule(r){9-11}
    Models & Natural & 0.1  & 0.2 & 0.3 & 0.1  & 0.2 & 0.3 & 0.1  & 0.2 & 0.3\\
    \midrule
    Natural Model & 99.1  & 72.7  & 28.1  & 7.3   & 22.5  & 0.0 & 0.0  & 21.6  & 0.0 & 0.0\\
    Xent-reg Model[6]  & 99.2  & \bf{91.7}  & 60.4  & 18.3   & 87.9  & 19.9 & 0.0  & 88.09  & 20.0 & 0.0\\
    Logit-reg Model  & 98.4  & 91.6  & \bf{68.5} & \bf{28.2}  & \bf{90.0}  & \bf{52.6} & \bf{5.9}  & \bf{92.2}  & \bf{69.4} & \bf{32.4}\\
    \bottomrule
  \end{tabular}
\end{table}

\begin{table}
  \caption{Test Accuracy of adversarial examples on CIFAR10 dataset (\%)}
  \label{table-cifar}
  \centering
  \begin{tabular}{lllllllllll}
    \toprule
    \multicolumn{3}{}{}     & FGSM   &&&    BIM       &&& C\&W   \\
    \cmidrule(r){3-5} \cmidrule(r){6-8} \cmidrule(r){9-11}
    Models & Natural & 3  & 6 & 9   & 3  & 6 & 9  & 3  & 6 & 9\\
    \midrule
    Natural Model & 75.3  & 6.2   & 2.6  & 1.6   & 0.1  & 0.0 & 0.0  & 0.0  & 0.0 & 0.0\\
    Xent-reg Model[6]  & 71.1  & 19.1  & 9.5  & 6.1   & 2.6  & 0.7 & 0.4  & 2.1  & 1.5 & 1.4\\
    Logit-reg Model  & \bf{78.8}  & \bf{53.7}  & \bf{36.5} & \bf{26.2}  & \bf{50.1}  & \bf{25.8} & \bf{17}  & \bf{48.2}  & \bf{25.6} & \bf{17.3}\\
    \bottomrule
  \end{tabular}
\end{table}

\subsection{Robust Training Performance on CIFAR10}

To evaluate the robustness of our proposed defense method on CIFAR10, we implement a five-layer neural network with three convolutional layers and two fully connected layers, which has 78.8\% accuracy.
Note that such a relatively low baseline is caused by the insufficient model capacity and will not impact the performance comparison results of the three models -- natural, Xent-reg, and Logit-reg model.
Different with the default experimental setting, the two iteration-based adversarial attacks -- BIM and C\&W are trained by 10 iterations with step size of 1 in this evaluation.
The $l_{\infty}$ is set to be 3, 6, and 9 respectively and corresponding results are obtained.
The evaluation and comparison results are demonstrated in Table.~\ref{table-cifar}.


The results indicate that the accuracy almost drops to 0 under the three attacks without any defense. 
Compared with the state-of-the art method -- Xent-reg, our proposed method can achieve about 4$\times$ accuracy under FGSM attack and all the three $l_{\infty}$ constraints.
Similar to the results on MNIST, the previous Xent-reg method cannot defend the BIM and C\&W under all the three $l_{\infty}$ constraints, e.g. only 2.6\%, 0.7\% and 0.4\% accuracy is obtained in the perspective of the BIM attack.
However, our proposed method can still achieve considerable accuracy and outperforms the Xent-reg as large as 40$\times$ against BIM attack and 20$\times$ against C\&W attack.

\begin{figure}[t]
  \centering
  \includegraphics[width=5.5in]{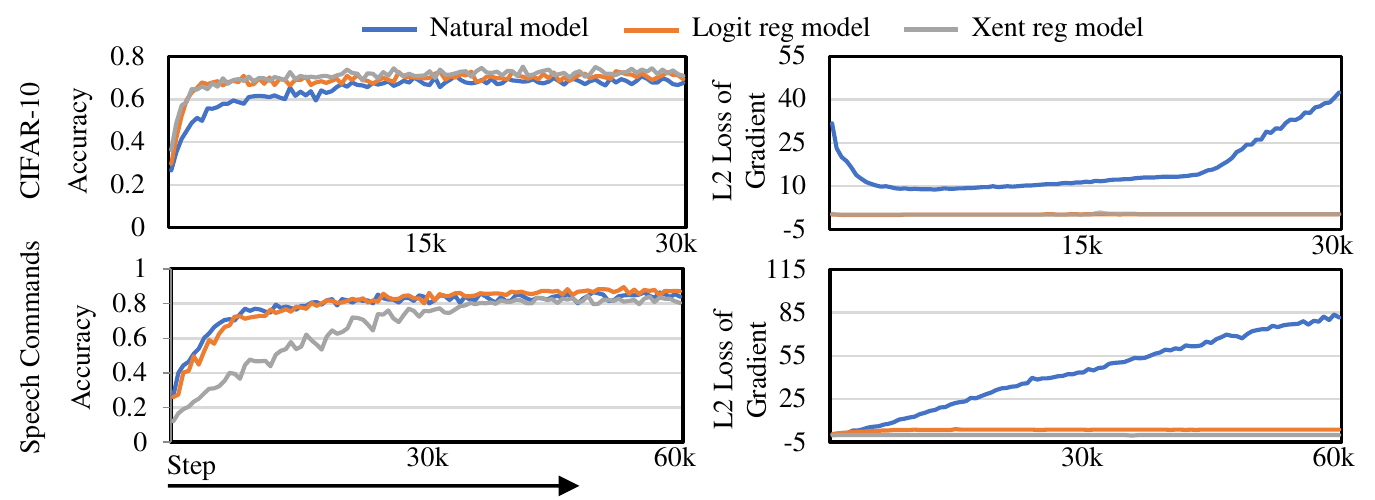}
  \caption{Accuracy and gradient loss curve during training process for CIFAR10 (first row) and Speech Command dataset (second row). As is shown, gradient regularizing effectively prevents large gradients' forming in neural network training process.}
  \label{fig:2}
\end{figure}

Fig.~\ref{fig:2} shows the test accuracy and gradient loss of the three models during their training process on CIFAR10 dataset (the above two figures in the first row).
Compared with natural model without any defense method, our gradient regularizing can constrain the $l_{2}$ loss of gradients within an extremely small range, which is around 0 and is reduced over 200 times. 
  To further evaluate the effectiveness of the proposed gradient regularizing, we randomly choose 1000 images from testing set and calculate their average $l_{1}$ norm of the Jacobian matrix on Natural Logit-reg models. 
  The experiment results show that the average gradients magnitude in Logit-reg models is 5$\times$ less than the gradients in the Natural model. 
  Such results indicate that larger magnitude perturbations are required to conduct effective attacking and prove robustness improvement of our proposed defense method. 
  

\subsection{Robust training performance on Speech Command Dataset}

Google Speech Command dataset [16] contains around 40000 one-second voice commands which labeled into twelve classes, e.g. yes, no, up, down, left, etc. 
  To classify the commands, we adopt a three-layer CNN model [17] with two convolutional layers with one fully connected layers. 
  Before feeding the audio waves into the neural network, we first pre-process the audio inputs and extract their 40-dimensional MFCC features. 
  These features are then fed into the neural network. 
  Our naturally trained model could achieve 87.0\% accuracy on test dataset. 
  Different from image inputs, in our adversarial examples generation process, all the adversarial perturbations are added to the MFCC features instead of the raw audios. This is because MFCC features are the direct input for neural networks. We implement same kind of attacks as before and then test models' robustness. The iteration-based adversarial example generation are under such settings: 8 iterations with step size of 0.002. The detailed results are shown in Table.~\ref{table-speech}.  

In audio dataset, our Logit-reg model also provides the highest accuracy under all adversarial settings. Especially, under all FGSM attacks, our model achieve 55.2\%, 31.7\% and 16.5\%, which are all 2 times higher than Xent-reg Model. This again shows our regularizer's advantage compared to cross-entropy regularizer, regardless of the type of classification tasks. Although both gradient regularizing methods incur small degrees of decrease on natural accuracy, logit regularizer shows less harm and meanwhile provide higher robustness against adversarial attacks. Fig.~\ref{fig:2}(b) shows the accuracy and gradient loss in the training process.

\begin{table}
  \caption{Test Accuracy of adversarial examples on Google Speech Command Dataset (\%)}
  \label{table-speech}
  \centering
  \begin{tabular}{lllllllllll}
    \toprule
    \multicolumn{3}{}{}     & FGSM   &&&    BIM       &&& C\&W   \\
    \cmidrule(r){3-5} \cmidrule(r){6-8} \cmidrule(r){9-11}
    Models  & Natural    & 0.25  & 0.5 & 0.75 & 0.25  & 0.5 & 0.75 & 0.25  & 0.5 & 0.75\\
    \midrule
    Natural Model & \bf{87.0}  & 17.0  & 9.6  & 9.1   & 11.5  & 0.1 & 0.0  & 11.5  & 0.0 & 0.0\\
    Xent-reg Model[6]  & 81.8  & 27.6  & 8.6  & 8.5   & 20.8  & 8.7 & 8.4  & 20.2  & 8.6 & 8.4\\
    Logit-reg Model  & 84.6  & \bf{55.2}  & \bf{31.7} & \bf{16.5}  & \bf{48.6}  & \bf{16.8} & \bf{8.9}  & \bf{47.9}  & \bf{17.0} & \bf{9.3}\\
    \bottomrule
  \end{tabular}
\end{table}

\begin{table}
  \caption{Defensive ability for transferable attacks on MNIST dataset (\%)}
  \label{table-mnist-trans}
  \centering
  \begin{tabular}{llllllllll}
    \toprule
    \multicolumn{2}{}{}     & FGSM   &&&    BIM       &&& C\&W   \\
    \cmidrule(r){2-4} \cmidrule(r){5-7} \cmidrule(r){8-10}
    Models  & Natural  & Xent & Logit & Natural  & Xent & Logit & Natural  & Xent & Logit\\
    \midrule
    Logit-reg   & \bf{85.8}  & \bf{55.6}  & 34.8   & \bf{79.0}  & \bf{42.8} & 11.0  & \bf{76.8}  & \bf{47.8} & \bf{47.4}\\
    Xent-reg[6]    & 76.8   & 19.8  & \bf{38.3}   &66.6   & 0.0 & \bf{15.8}  &69.0   & 14.0 & 43.0\\
    \bottomrule
  \end{tabular}
\end{table}

\begin{table}
  \caption{Defensive ability for transferable attacks on CIFAR10 dataset (\%)}
  \label{table-cifar-trans}
  \centering
  \begin{tabular}{llllllllll}
    \toprule
    \multicolumn{2}{}{}     & FGSM   &&&    BIM       &&& C\&W   \\
    \cmidrule(r){2-4} \cmidrule(r){5-7} \cmidrule(r){8-10}
    Models  & Natural  & Xent & Logit & Natural  & Xent & Logit & Natural  & Xent & Logit\\
    \midrule
    Logit-reg  & \bf{81.0}  &\bf{81.0} & 37.0  & \bf{83.0}  & \bf{81.0}  & 30.0  & \bf{82.0}  & \bf{81.0}  & 31.0\\
    Xent-reg[6]  & 36.0  & 12.0 & \bf{44.0}  & 57.0  & 3.0  & \bf{41.0}  & 56.0  & 6.0  & \bf{42.0}\\
    \bottomrule
  \end{tabular}
\end{table}

\begin{table}
  \caption{Defensive ability for transferable attacks on Google Speech Command dataset (\%)}
  \label{table-speech-trans}
  \centering
  \begin{tabular}{llllllllll}
    \toprule
    \multicolumn{2}{}{}     & FGSM   &&&    BIM       &&& C\&W   \\
    \cmidrule(r){2-4} \cmidrule(r){5-7} \cmidrule(r){8-10}
    Models  & Natural  & Xent & Logit & Natural  & Xent & Logit & Natural  & Xent & Logit\\
    \midrule
    Logit-reg   & \bf{37.3}  & \bf{39.9}  & 16.5   & \bf{31.1}  & \bf{33.0} & 8.9  & \bf{31.9}  & \bf{35.3} & 9.3\\
    Xent-reg[6]    & 18.9  & 10.8  & \bf{19.5}   & 13.7  & 8.4 & \bf{14.2}  & 13.9  & 8.4 & \bf{16.7}\\
    \bottomrule
  \end{tabular}
\end{table}

\subsection{Robust Training Performance Against Transferable Attack}

Besides white-box adversarial attacks, we further evaluate our robust trained models against transferable attacks. 
On MNIST, CIFAR10 and Google Speech Command datasets, we use FGSM, BIM and C\&W methods to generate adversarial examples on Natural, Logit-reg and Xent-reg models separately.  Note the models we used for adversarial examples generation are totally different from our test models. Thus, such generated adversarial example are used as transferable adversarial examples. 
Then we feed these adversarial examples into Logit-reg and Xent-reg models to evaluate their defensive ability for transferable attacks. 
The results are shown in Table.~\ref{table-mnist-trans}, Table.~\ref{table-cifar-trans} and Table.~\ref{table-speech-trans}.
For all the three attack methods, we use strongest attack settings. Therefore the $l_{\infty}$ is set to 0.3, 9 and 0.75 in MNIST, CIFAR10 and Google Speech Command respectively. 

Among all the datasets and attacks, Logit-reg model always achieves much higher accuracy for natural and Xent-reg models' generated adversarial examples from 33\% to 81\%. This shows that Logit-reg model are more robust than Xent-reg models when facing natural and Xent-reg models' adversarial examples. 
As for the Logit-reg model based adversarial examples, it cannot beat the Xent-reg model's accuracy. We suppose that this is because the two Logit-reg models have learned the similar decision boundaries. Moreover, from the Table.~\ref{table-mnist-trans}, we can see Logit-reg model gets 55.6\%, 42.8\% and 47.8\% accuracy rate for adversarial examples which generated by Xent-reg model for three attack methods. In the contrast, Xent-reg model can only get 38.3\%, 15.8\% and 43.0\% accuracy rate for adversarial examples generated from Logit-reg model. This difference also existed in both Table.~\ref{table-cifar-trans} and Table.~\ref{table-speech-trans}. So, our Logit-reg model has strong defensive ability for transferable attacks than Xent-reg model in all the three dataset. 

\section{Discussion}
\begin{figure}[t]
  \centering
  \includegraphics[width=5.5in]{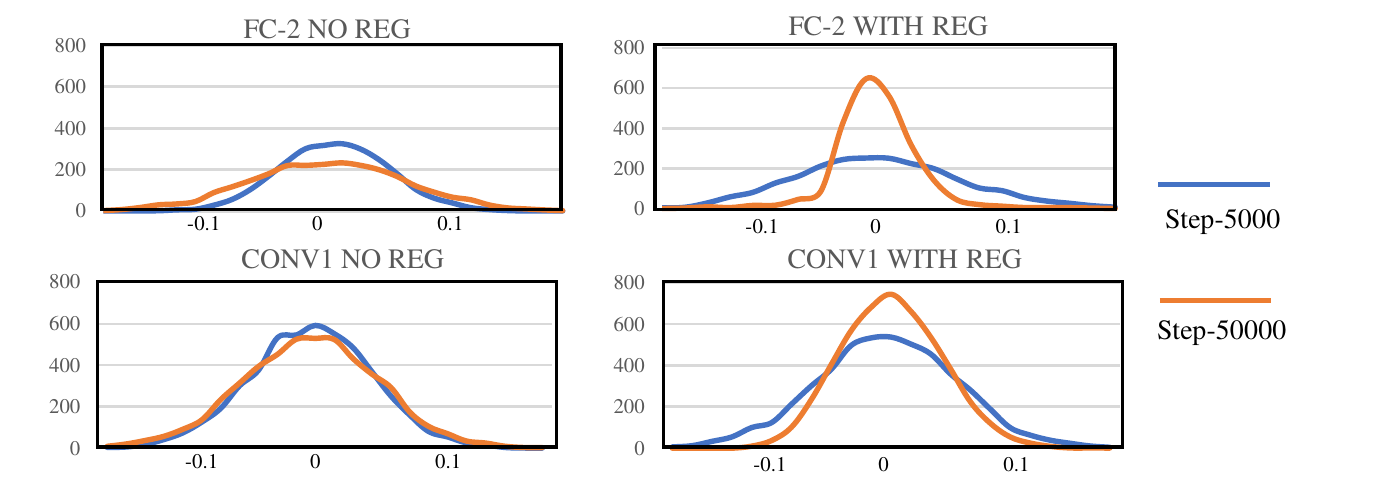}
  \caption{Weights distribution change in convolution layer-1 and fully connected layer-2 during normal model training (left column) and robust training process (right column). Step 5000 and 50000 are chosen in the training starting stage and final converging stage, respectively. Through comparison, we could see that gradient regularizing imposes penalty on all the weights to form a relatively "sharp" distribution around 0 point, which means large amounts of weights are becoming nearer 0.}
  \label{fig:3}
\end{figure}

\paragraph{Weight Distribution Squeezing Analysis}
In our experiment, we find that our logit regularizer could not only effectively regulate the gradients of neural networks, but can also be able to regulate the weights distributions towards to its minimal variance and mean value. As illustrated in Fig.~\ref{fig:3}, the left two sub-figures represent the neutral network layer without adding our logit regularizer while right two sub-figures represent the layer with our regularizer. For left figures, the weights distributions in both CONV1 and FC-2 do not have significant change. In the contrast, for right figures, after 10 times steps, the weights has been squeezed to a smaller mean and variance value. We suppose this is because the gradient regulation process finally forces each element in the weight matrix to be smaller towards lower gradients loss. 

\paragraph{Compare to State-of-Art Defense}
In this part, we compare our gradient regularizing training method with the state-of-art robustness training method: Min-Max training [15] by considering the time consumption.
	To guarantee the model robustness, Min-Max training needs to generate over ten times of adversarial examples than the given batch size during every step of training. 
	This process will significantly increase the total training time and limits Min-Max training to only some small datasets and cannot generalize to real-word big datasets, like ImageNet[7]. 
	By contrast, our method can improve the network models' robustness directly in the training process, with only double-backpropagation overhead.

To give a brief view, we evaluate the time consumption of two robustness training methods on CIFAR10. For same 10 thousand steps training (each step train 300 samples), our method only cost 12 minutes while Min-Max used more than 1 hour and 16 minutes, which is a 6 times time difference.    
\section{Conclusion}

In this work, we re-think the adversarial examples and neural networks in a \textit{quasi-linear} view. 
	Based on the insight that first-order gradient accounts for the main output change w.r.t unit perturbation $\Delta x$, we propose a novel robust training method by regulating the first-order gradients of neural networks. 
	Further, we design a optimal regularizer to effectively capture the adversarial gradients and then regulate them by double-backpropagation in the training process. 
Experiments show that our method could greatly enhance model's robustness against a large array of adversarial examples and outperform the state-of-art gradient regularizing method with relatively low overhead.

\newpage
\section*{References}
\small



[1] Goodfellow, I.J., Shlens, J. \ \& Szegedy, C.\ (2017) Explaining and harnessing 
adversarial examples. {\it arXiv preprint arXiv:1412.6572}.

[2] Szegedy, C., Zaremba, W., Sutskever, I., Bruna, J., Erhan, D., Goodfellow,
 I.\ \& Fergus, Rob\ (2013) Intriguing properties of neural networks. {\it arXiv preprint arXiv:1312.6199}.

[3] LeCun, Y. \ \& Bengio, Y.\ (1995) Convolutional networks for images, speech, and time series. {\it The handbook of brain theory and neural networks}.

[4] Drucker, H.\ \& Le C.Y.\ (1992) Double backpropagation increasing generalization
 performance. In {\it IJCNN}, pp.\ 145-150. 

[5] Carlini, N.\ \& and Wagner, D.\ (2017) Towards evaluating the robustness
 of neural networks. In {\it Security and Privacy (SP), 2017 IEEE Symposium}, 
 pp.\ 39-57.

[6] Ross, A.S. \ \& Doshi-Velez, F.\ (2017). Improving the Adversarial Robustness and Interpretability of Deep 
Neural Networks by Regularizing their Input Gradients. In {\it arXiv preprint arXiv:1711.09404}.

[7] Krizhevsky, A., Sutskever, I. \ \& Hinton, G.E.\ (2012) Imagenet classification 
with deep convolutional neural networks. In {\it Advanced in Neural Information
 Processing System}, pp.\ 1097-1105.

[8] Simonyan, K. \ \& Zisserman, A.\ (2014) Very deep convolutional networks for 
large-scale image recognition. {\it arXiv preprint arXiv:1409.1556}.

[9] Papernot, N., McDaniel, P., Jha, S., Fredrikson, M., Celik, Z. \ \& Swami, A.\ (2016) The
 limitations of deep learning in adversarial settings. In {\it Security and Privacy (EuroS\&P), IEEE European Symposium}, 
 pp.\ 372-387. 

[10] Papernot, N., McDaniel, P. \ \& Goodfellow, I.\ (2016) Transferability in machine learning: from phenomena to black-box attacks using adversarial samples. In {\it arXiv preprint arXiv:1605.07277}.

[11] Liu, Y., Chen, X., Liu, C. \ \& Song, D.\ (2016) Delving into transferable adversarial examples and black-box attacks. In {\it arXiv preprint arXiv:1611.02770}.

[12] Papernot, N., McDaniel, P., Goodfellow, I., Jha, S., Celik, Z.B. \ \& Swami, A.\ (2017) Practical
 black-box attacks against machine learning. In {\it Proceedings of the 2017 ACM on Asia Conference on Computer and Communications Security}, 
 pp.\ 506-519.

[13] Papernot, N., McDaniel, P., Wu, X., Jha, S. \ \& Swami, A.\ (2016) Distillation
 as a defense to adversarial perturbations against deep neural networks. In {\it Security and 
 Privacy (SP), 2016 IEEE Symposium}, pp.\ 582-597.

[14] Papernot, N. \ \& McDaniel, P.\ (2017) Extending Defensive Distillation. In {\it arXiv preprint arXiv:1705.05264}.

[15] Madry, A., Makelov, A., Schmidt, L., Tsipras, D. \ \& Vladu, A.\ (2017) Towards deep 
learning models resistant to adversarial attacks. {\it arXiv preprint arXiv:1706.06083}.

[16] P. Warden.\ (2017) Speech commands: A public dataset for single-word speech recognition. {\it Dataset available}.

[17] Sainath, T.N. \ \& Parada, C.\ (2015) Convolutional neural networks for small-footprint keyword spotting. In {\it Sixteenth Annual Conference of the International Speech Communication Association}.



\end{document}